\documentclass[preprint,12pt,3p]{elsarticle}

\usepackage{graphicx}
\usepackage{amssymb}
\usepackage{subfig}
\usepackage{amsmath}

\begin{document}

\begin{frontmatter}
\title{Augmented Reality for Depth Cues in Monocular Minimally Invasive Surgery}

\author[label1]{Long Chen}
\address[label1]{Bournemouth University}
\ead{chenl@bournemouth.ac.uk}

\author[label1]{Wen Tang}
\ead{wtang@bournemouth.ac.uk}

\author[label2]{Nigel W. John}
\address[label2]{University of Chester}
\ead{nigel.john@chester.ac.uk}

\author[label3]{Tao Ruan Wan}
\address[label3]{University of Bradford}
\ead{t.wan@bradford.ac.uk}

\author[label1]{Jian Jun Zhang}
\ead{jzhang@bournemouth.ac.uk}

\begin{abstract} One of the major challenges in Minimally Invasive Surgery (MIS) such as laparoscopy is the lack of depth perception. In recent years, laparoscopic scene tracking and surface reconstruction has been a focus of investigation to provide rich additional information to aid the surgical process and compensate for the depth perception issue. However, robust 3D surface reconstruction and augmented reality with depth perception on the reconstructed scene are yet to be reported. This paper presents our work in this area. First, we adopt a state-of-the-art visual simultaneous localization and mapping (SLAM) framework - ORB--SLAM - and extend the algorithm for use in MIS scenes for reliable endoscopic camera tracking and salient point mapping. We then develop a  robust global 3D surface reconstruction framework based on the sparse point clouds extracted from the SLAM framework. Our approach is to combine an outlier removal filter within a Moving Least Squares smoothing algorithm and then employ Poisson surface reconstruction to obtain smooth surfaces from the unstructured sparse point cloud. Our proposed method has been quantitatively evaluated compared with ground-truth camera trajectories and the organ model surface we used to render the synthetic simulation videos. \textit{In vivo} laparoscopic videos used in the tests have demonstrated the robustness and accuracy of our proposed framework on both camera tracking and surface reconstruction, illustrating the potential of our algorithm for depth augmentation and depth-correct augmented reality in MIS with monocular endoscopes.
 
\end{abstract}


\end{frontmatter}

\section{Introduction}
\label{intro}
Minimally Invasive Surgery (MIS) requires a surgeon to successfully perform technically demanding procedures.  Typically their visual interface consists of a monocular display showing the video stream from the endoscope (although there are some examples of stereoscopic endoscopes discussed below). The Field of View (FOV) captured by a monocular endoscopic camera is usually limited, for example, only 30\% to 40\% of the whole liver surface is visible in one frame \cite{Plantefeve2016}. Further, the loss of depth perception from monocular image sequences can severely impact on a surgeon's performance in completing complex tasks \cite{Honeck2012} \cite{Wagner2012}. As such, providing cues to aid better depth perception is highly desirable for a monocular endoscope scene and requires further innovation. In this paper we propose a novel method of fusing rich 3D anatomical information with a monocular endoscope video stream, based on real-time 3D reconstruction of the scene to address the limited FOV and depth perception issue. There are three main technical challenges to overcome: (i) achieving real-time tracking and feature extraction of the scene; (ii) dealing with camera tracking and the sparse and noisy data set extracted from the camera tracking algorithm; and (iii) robust reconstruction of a 3D surface of the scene. 


Recent advances in computer hardware and software technologies has facilitated the use of computer vision techniques for MIS scene analysis and understanding. For example, efficient motion tracking of endoscopic cameras in MIS for deformable soft tissues has been demonstrated  \cite{MountneyLoThiemjarusEtAl2007}, as well as augmented reality with anatomy structures \cite{HaouchineDequidtPeterlikEtAl2013} \cite{HaouchineCotinPeterlikEtAl2015}. There are particular challenges, however, since in MIS the luminance changes dramatically and an endoscope can also move rapidly during insertion and extrusion. The scale-invariant feature transform (SIFT) algorithm \cite{Kim2012} and Speeded Up Robust Features (SURF) algorithm \cite{Kumar2014} have been developed for robust feature based camera motion tracking, and other approaches specifically designed to work with soft tissues that account for scale, rotation and brightness \cite{MountneyYang2008}. However, the issue of depth perception remains regardless of whichever computer vision algorithm has been deployed, i.e. information regarding the depth of elements within a scene has not been recovered and selected feature points extracted from vision algorithms must be within the field of view to provide the geometrical information required by augmented reality.

A stereo endoscope will improve the depth perception problem, and there are such systems available - often integrated into robotic systems (e.g. the da Vinci system from Intuitive Surgical, Inc.) or using proprietary stereo cameras. 3D depth can be recovered using the disparity map from rectified stereo images during a laparoscopic surgery \cite{StoyanovDarziYang2004} \cite{Stoyanov2005}, so that dense 3D reconstruction of the laparoscopic scene can be achieved by a propagation method \cite{StoyanovScarzanellaPrattEtAl2010} and/or a cost-volume algorithm \cite{ChangStoyanovDavisonEtAl2013}. Stereo vision based reconstructions, however, can only recover the structure of a local frame without a global overview of the scene. Therefore, such local reconstruction is prone to tracking errors and tracking noise. In addition, stereo endoscopic surgery is still expensive and yet to be used widely in practice, compared with the monocular endoscope.

Recently, the maturity of the simultaneous localization and mapping (SLAM) method used by robot devices for navigation in an unknown 3D environment has opened up new opportunities for novel camera tracking approaches in MIS. A SLAM-enabled system makes it possible to estimate the 3D structure of an unknown environment from a moving camera and simultaneously track the pose of the camera in the environment. The scenario of tracking and scene reconstruction in endoscopic surgery is similar to that of a typical SLAM application. Further, tracking and mapping using SLAM can be very accurate (within 1mm in ideal conditions) and does not require the use of optical or magnetic trackers that may distract a surgeons' view. EKF-based SLAM has already been widely used with laparoscopic image sequences \cite{Mountney2006} \cite{Mountney2009} \cite{GrasaBernalCasadoEtAl2014}. Although the further motion compensation model \cite{MountneyYang2010} and stereo semi-dense reconstruction \cite{Totz2011} are integrated into the EKF-SLAM framework, due to the linearization of motion and sensor models by first-order Taylor series expansion, the accuracy of EKF-SLAM cannot be guaranteed and therefore it is prone to inconsistent estimation and drifting. The first keyframe-based SLAM -- PTAM (Parallel Tracking and Mapping) was a breakthrough in visual SLAM and has also been used in MIS for simultaneous stereoscope tracking \cite{Lin2013}. ORB--SLAM is a well-designed SLAM system derived from the idea of PTAM, which utilizes ORB (Oriented FAST and Rotated BRIEF) binary features for fast and reliable feature point tracking. Mahmoud \textit{et al} \cite{NaderMahmoud2016} tested ORB--SLAM on endoscope video and presented a method for map point densifying but with some loss of accuracy.

In this paper, we adapt ORB--SLAM for endoscopic camera tracking and mapping. A 3D surface reconstruction method is proposed based on Moving Least Squares (MLS) smoothing and Poisson surface reconstruction to recover a smooth surface from the  unstructured sparse map points extracted from ORB--SLAM. We also provide a comprehensive quantitative assessment by using simulated laparoscopic sequences. Estimated camera trajectories and reconstructed surfaces are compared with ground truth camera trajectories and the 3D models that we used to render the simulated video. The experiment results yield root mean square errors (RMSE) of 1.24 mm for camera trajectories and 4.32mm for surface reconstruction. Our method provides new possibilities for depth augmentation in monocular endoscopic MIS, enabling surgeons to view correct depth during a procedure. We also demonstrate an augmented reality (AR) framework for superimposing virtual objects with the correct depth, which is very important to prevent object drifting when changing perspectives.

\section{ORB--SLAM for endoscopic camera tracking and mapping}
\label{method}

\begin{figure}
\centering
\includegraphics[width=0.75\textwidth]{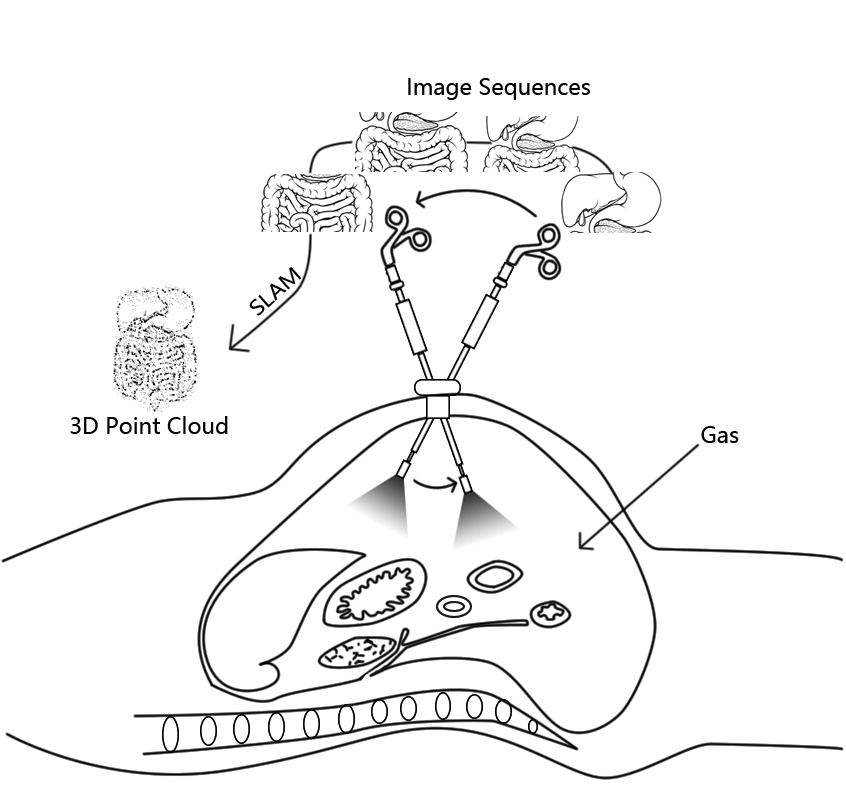}
\caption{A moving monocular endoscopic camera can capture a series of image sequences which can be used to build a 3D sparse point cloud by using a SLAM system.}
\label{fig}       
\end{figure}

 ORB--SLAM \cite{Mur-ArtalMontielTardos2015} combines many state-of-the-art techniques into one SLAM system, such as using an ORB descriptor for tracking, local keyframe for mapping, graph-based optimization, Bag of Words algorithm for relocalization, and an essential graph for loop closure. These features can enable real-time endoscopic camera tracking and sparse point mapping in an abdominal cavity as shown in Fig. \ref{fig}. Real-time performance is crucial in time-demanding medical interventions. Since ORB is a binary feature point descriptor, it is an order of magnitude faster than SURF and more than two orders faster than SIFT and also offers better accuracy. In addition, ORB features are invariant to rotation, illumination and scale, which means that it is capable of dealing with the rapid movements of endoscope cameras including rotation, zooming and change in brightness.


A common problem for monocular scene analysis using SLAM is initialized, i.e. a procedure is required to create an initial map, as depth cannot be recovered from a single image. An automatic approach is used in ORB--SLAM to calculate homography for planar scenes and a fundamental matrix for non-planar scenes dynamically. This approach can greatly increase the success rate of initialization and reduce the initializing time, which also facilitates its use in a MIS scene for initialization on an organ surface or to compute a fundamental matrix when the endoscopic camera is pointing at a complex structure.

The use of the Bags of Words (BoW) algorithm in ORB--SLAM can also help relocalization when tracking is lost. The vocabulary is created offline with a large number of ORB descriptors extracted from very big datasets with indoor and outdoor images, which almost covers all of patch patterns that we can encounter. The vocabulary serves as a classifier or a dictionary that assigns each descriptor an index. When a new image enters the system, each descriptor of features in this image is looked up, and a unique vector will be built based on the index of descriptors. In this way, the rough similarity of two images can be acquired by simply comparing the two unique vectors, which can greatly increase the speed of relocalization. For endoscopic videos, however, colours of soft tissues, organs and vessels do not differ greatly from person to person. Therefore, to extend the ORB--SLAM to be used in MIS scenes, we trained our vocabulary list specifically for its use in MIS based on a database of 7,894 images from the Hamlyn Centre Endoscopic Video Datasets \cite{London2016} \cite{Ye2016}. By using a specific BoW database, the length of unique vector for similarity measurement will be decreased, which could increase not only the speed for performing comparisons but also the accuracy.

To optimise for MIS scenes, we followed the method of \cite{NaderMahmoud2016} to fine tune some of the default parameters that were used in ORB--SLAM. We extend the searching region by a factor of 1.5 to allow more key points to be included. The parallax threshold for point initialization has also been increased by a factor of 5 to increase accuracy when triangulating the points for 3D positions. The maximum threshold that is allowed between keypoints and reprojected map points for triangulation is reduced by a factor of 10 to strictly select robust 3D points. Finally, Hamming distance threshold for ORB descriptors comparison is decreased by 0.8 for more strict application of the pair point rule.
\section{Intra-operative 3D surface reconstruction}

\begin{figure}
\captionsetup[subfigure]{labelformat=empty}
\subfloat[]{\includegraphics[width=1\textwidth]{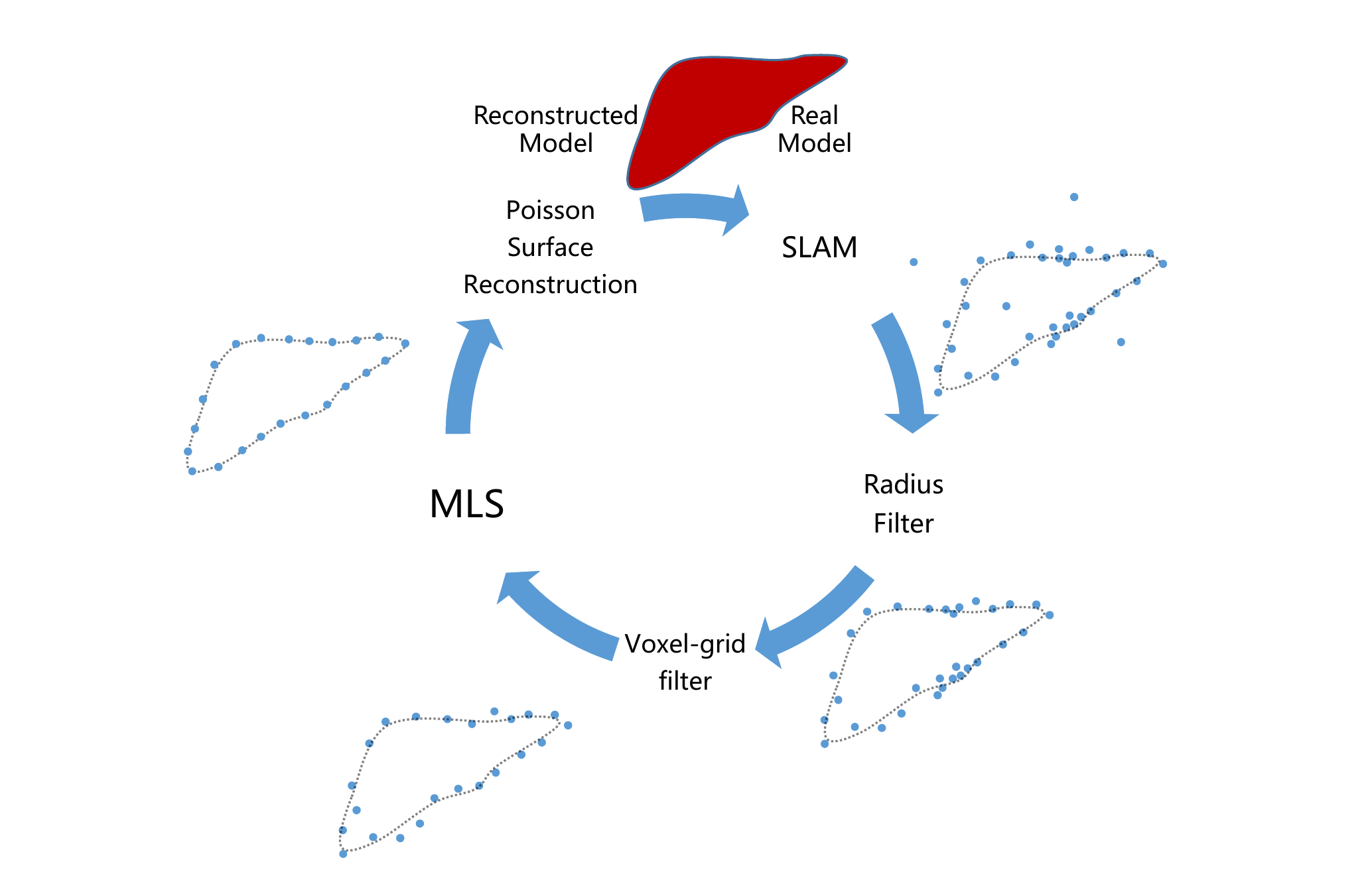}}\\
\subfloat[]{\includegraphics[width=1\textwidth]{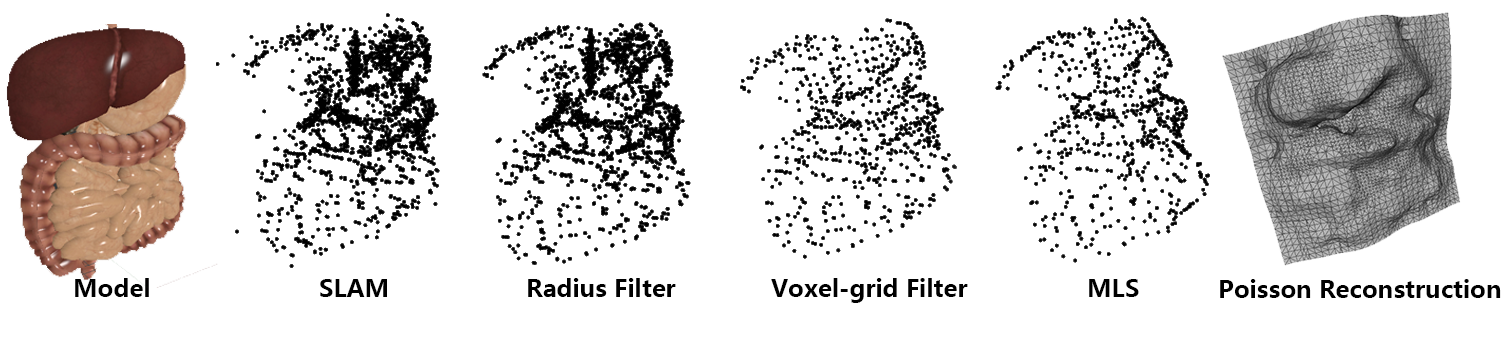}}\\
\caption{The proposed intra-operative 3D surface reconstruction framework.}
\label{recon}       
\end{figure}

In our system, the use of a SLAM system has made it possible to extract a sparse 3D point cloud from a moving monocular endoscopic camera. Unstructured sparse point clouds, however, describe the 3D structure of the endoscopic scene poorly. Therefore, we propose a 3D surface reconstruction framework that combines outlier removal filters, the Moving Least Square algorithm to smooth noise data and a Poisson surface reconstruction method to generate a dense and smooth surface from a unstructured sparse point cloud. This pipeline is illustrated in Fig. \ref{recon}.

\subsection{Point cloud pre-process}

The point cloud P given by ORB--SLAM represents the salient points that are visible at different camera keyframes, giving a sparse representation of the intra-operative scene. MIS scenes are very complex due to camera calibration, tissue movement and reflection, and so results in a noisy point cloud mixed with many outliers that affect the final surface reconstruction. Therefore, before feeding the point cloud into the reconstruction pipeline, we apply two outlier removal filters to remove the noisy points located amongst the raw data points. 

We first employ a radius filter that processes points in a cloud based on the number of neighbours that the points have. Points with very few neighbours are labelled as outliers, as isolated points cannot describe the structure of the 3D scene. Since some texture-abundant areas gain many more points than other areas, a voxel-grid filter is then used to re-sample the point cloud into a more even point cloud. After the filtering process, the point cloud becomes flat and ready for MLS smoothing and 3D surface reconstruction. 


\subsection{Moving Least Squares for point smoothing}

The Moving Least Squares (MLS) algorithm \cite{Levin2004} reconstructs surfaces locally by solving an optimization problem to find a local reference plane and fit a polynomial to the surface. Let a point set $p_{i}\in \mathbb{R}_{3},i\in \left \{ 1,...,N \right \}$ be the point cloud produced from the ORB--SLAM system. The continuous and smooth MLS surface $S$ can be computed by a two-step procedure: (i) a local reference plane is defined as $H = \left \{ x \in \mathbb{R}_{3}|x\cdot n - D = 0 \right \}$, which can be computed by minimizing the weighted sum of squared distances: 
$$\sum_{i=1}^{n} (p_{i}\cdot n - D)^{2} \Phi (\left \| p_{i}-q \right \|)$$ 
where $q$ is the projection of $p$ onto $H$, and $\Phi$ is the MLS kernel, usually a Gaussian; (ii) after the points are projected onto the initial local reference plane, a second least squares optimization is used to find a bi--variate polynomial function $g(u,v)$ (where $u,v$ is the local coordinate of $q$ in $H$) that approximates to the local surface. The projection of $p$ onto $S$ can then be defined by the polynomial value at the origin, i.e. $q + g(0, 0)\cdot n$.

\subsection{Poisson surface reconstruction}

We represent the points after the MLS filter stage by a vector field $\overrightarrow{V}$. Poisson surface reconstruction \cite{Michael2006} approaches the surface reconstruction problem through a framework of implicit functions that compute a 3D indicator function $\chi$ (which is equal to 1 inside the model and 0 at the outside points). Therefore, the problem becomes finding the $\chi $ whose gradient can best approximate the vector filed $\overrightarrow{V}$: 
$$min_{\chi }\left \| \bigtriangledown \chi -\overrightarrow{V} \right \|$$
Applying the divergence operator, we can transform this into a Poisson problem:
$$\bigtriangledown \times (\bigtriangledown_{\chi }) = \bigtriangledown \times \overrightarrow{V} \Leftrightarrow \bigtriangleup \chi = \bigtriangledown \times \overrightarrow{V}$$
After solving the Poisson problem and obtaining the 3D indicator function $\chi$, the 3D surface can be directly obtained by extracting an isosurface. Poisson reconstruction acts as a global solution that treats all of the data points simultaneously without relying on heuristic partitioning or blending, so that it can robustly approximate noisy data and create very smooth surfaces.

\section{Results and Discussion}
\label{studies}
We designed a two-part quantitative and qualitative evaluation process: (i) using a realistic simulation of a MIS scene video for the ground truth study to assess the performance of ORB--SLAM tracking error and the accuracy of the proposed surface reconstruction framework; (ii) using a real \textit{in vivo} video acquired from the Hamlyn Centre Laparoscopic/Endoscopic Video Datasets \cite{London2016} \cite{Mountney2010} to assess the quality of two applications of our proposed framework i.e. depth augmentation and AR with correct depth.

\subsection{System setup}

Our system is implemented in an Ubuntu 14.04 environment using C/C++ (without any GPU acceleration). All experiments are conducted on a workstation equipped with Intel Xeon(R) 2.8 GHz quad core CPU, 32G Memory, and one NVIDIA GeForce GTX 970 graphics card. The size of the simulation image sequences is 1024 X 768 pixels and the size of \textit{in vivo} endoscope video is 840 X 640 pixels. ORB--SLAM with our proposed AR framework runs in real-time at 40 FPS at max and the 3D surface reconstruction process takes around 600ms to traverse the whole pipeline.

\subsection{Ground truth study using simulation data}

For the evaluation of the tracking performance in terms of tracking accuracy, camera trajectories estimated by ORB--SLAM were aligned with trajectories of the ground truth camera used to render the MIS scene video. Similarly, the accuracy of our proposed 3D surface reconstruction framework is evaluated by comparing the reconstructed surface with the 3D model used to render the simulation video.

\begin{figure}
\centering
\subfloat[]{\includegraphics[width=0.358\textwidth]{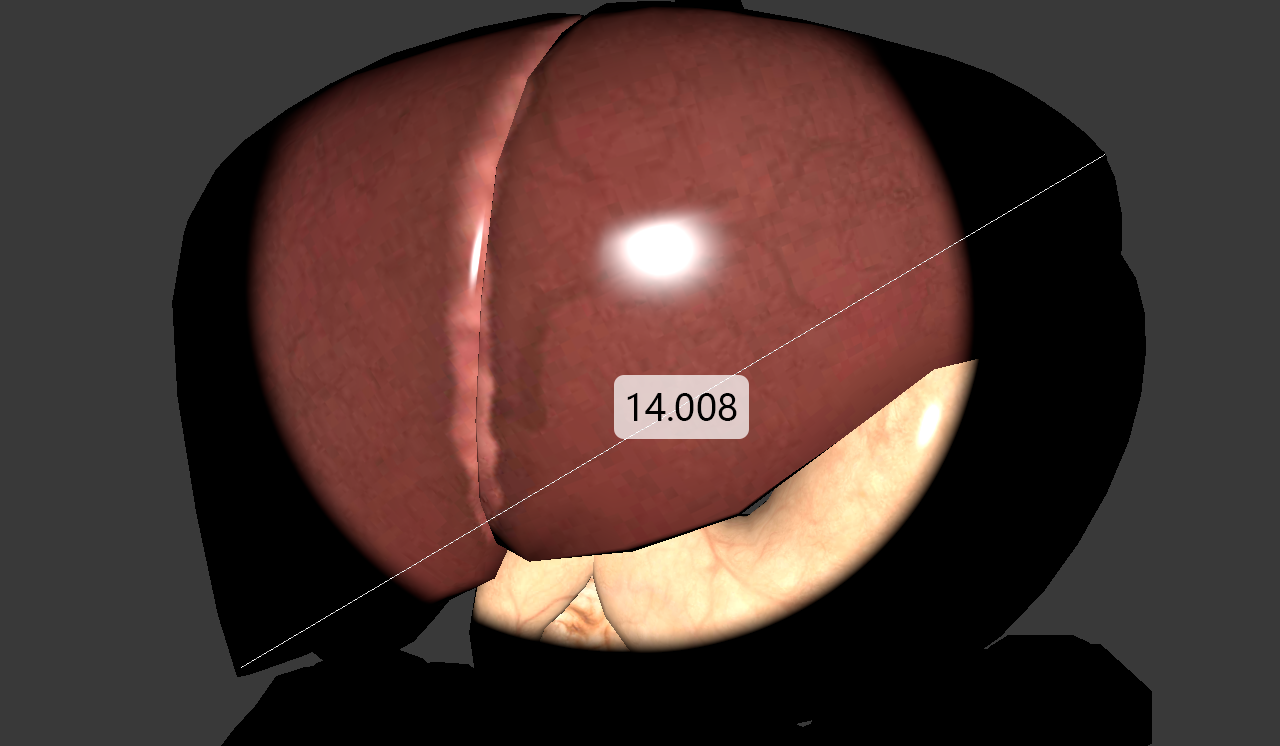}}\
\subfloat[]{\includegraphics[width=0.348\textwidth]{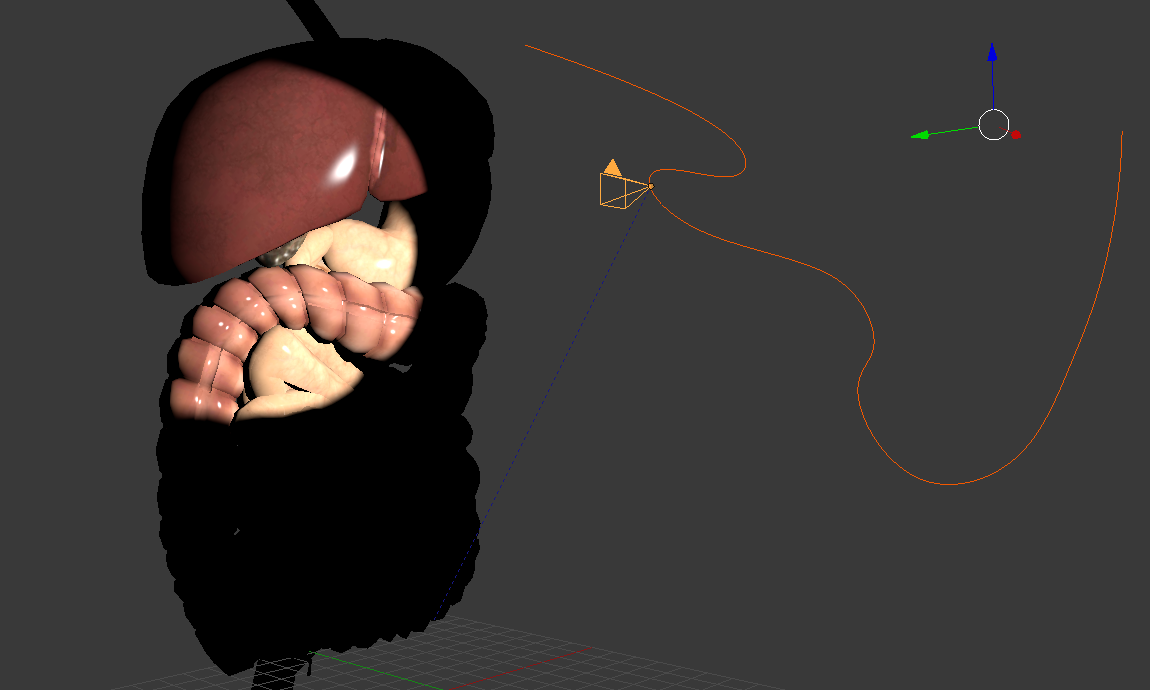}}\
\subfloat[]{\includegraphics[width=0.279\textwidth]{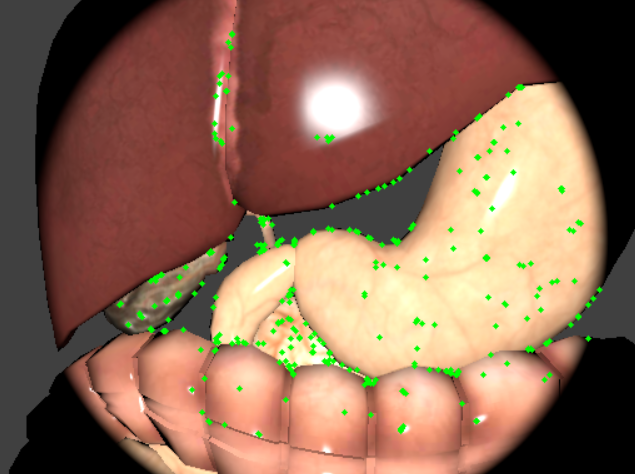}}\\
\caption{Simulated MIS scenes with a realistic human digestive system model. (a) The size of the model is scaled to the real world size of the size of an adult liver. (b) The only light is attached to the camera and camera trajectory is designed to hover around the 3D model. (c) The frame that ORB--SLAM succeeded in initializing.}
\label{size}       
\end{figure}

To quantitatively evaluate the performance of ORB--SLAM, we used the Blender \cite{Blender2016} -- an open source 3D creation software to render realistic image sequences of a simulated abdominal cavity scene using pre-defined endoscopic camera movement. The digestive system contains models with appropriate textures to make the scene as realistic as possible. The model was scaled to be real life size according to an average measured liver diameter of 14.0 cm \cite{Kratzer2003} as shown in Fig. \ref{size}(a); the material property was set with a strong specularity component to simulate smooth and reflective liver surface tissue. The luminance is intentionally set high by using a spot light attached to the main camera to simulate an endoscope camera as shown in Fig. \ref{size}(a) to render a realistic endoscopic lighting condition. We designed a camera trajectory that hovers around the 3D model as shown in Fig. \ref{size}(b) to capture as much area as possible for building a point cloud that covers the whole front surface of the models. There are 900 frames of image sequences at a frame-rate of 30 fps rendered out, which is equivalent to a 30 seconds video.

\begin{figure}
\centering
\subfloat[]{\includegraphics[width=0.45\textwidth]{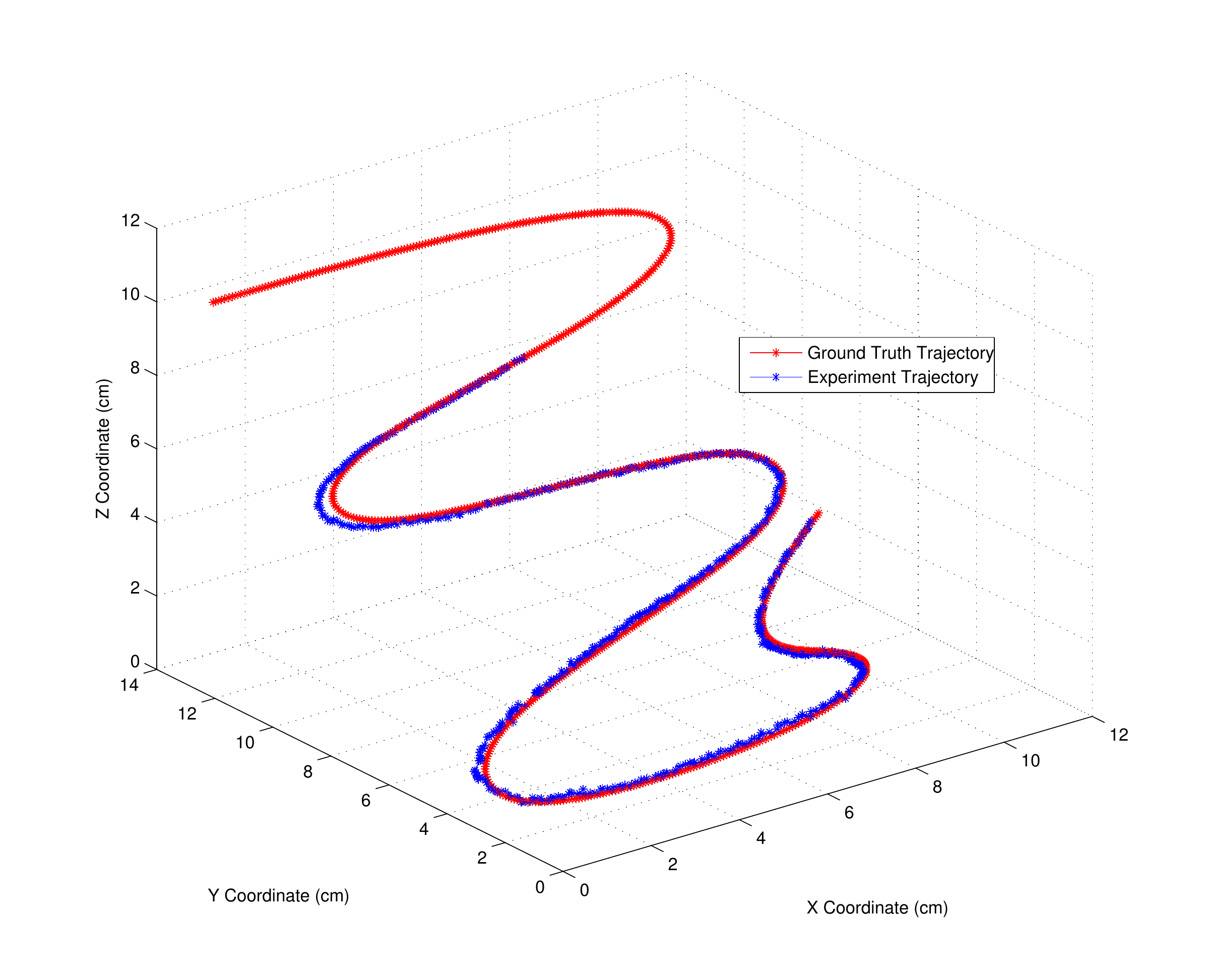}}\
\subfloat[]{\includegraphics[width=0.45\textwidth]{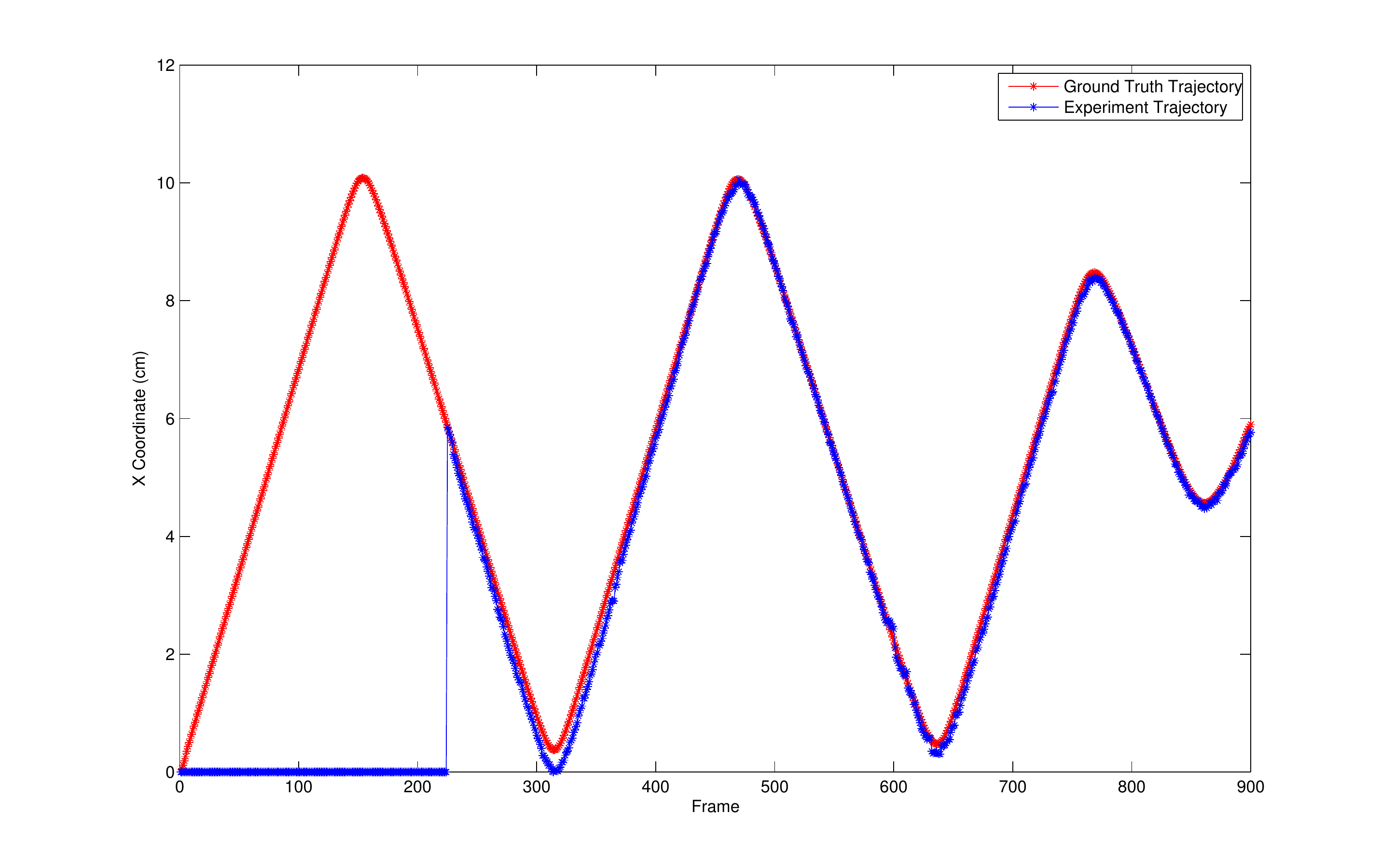}}\\
\subfloat[]{\includegraphics[width=0.45\textwidth]{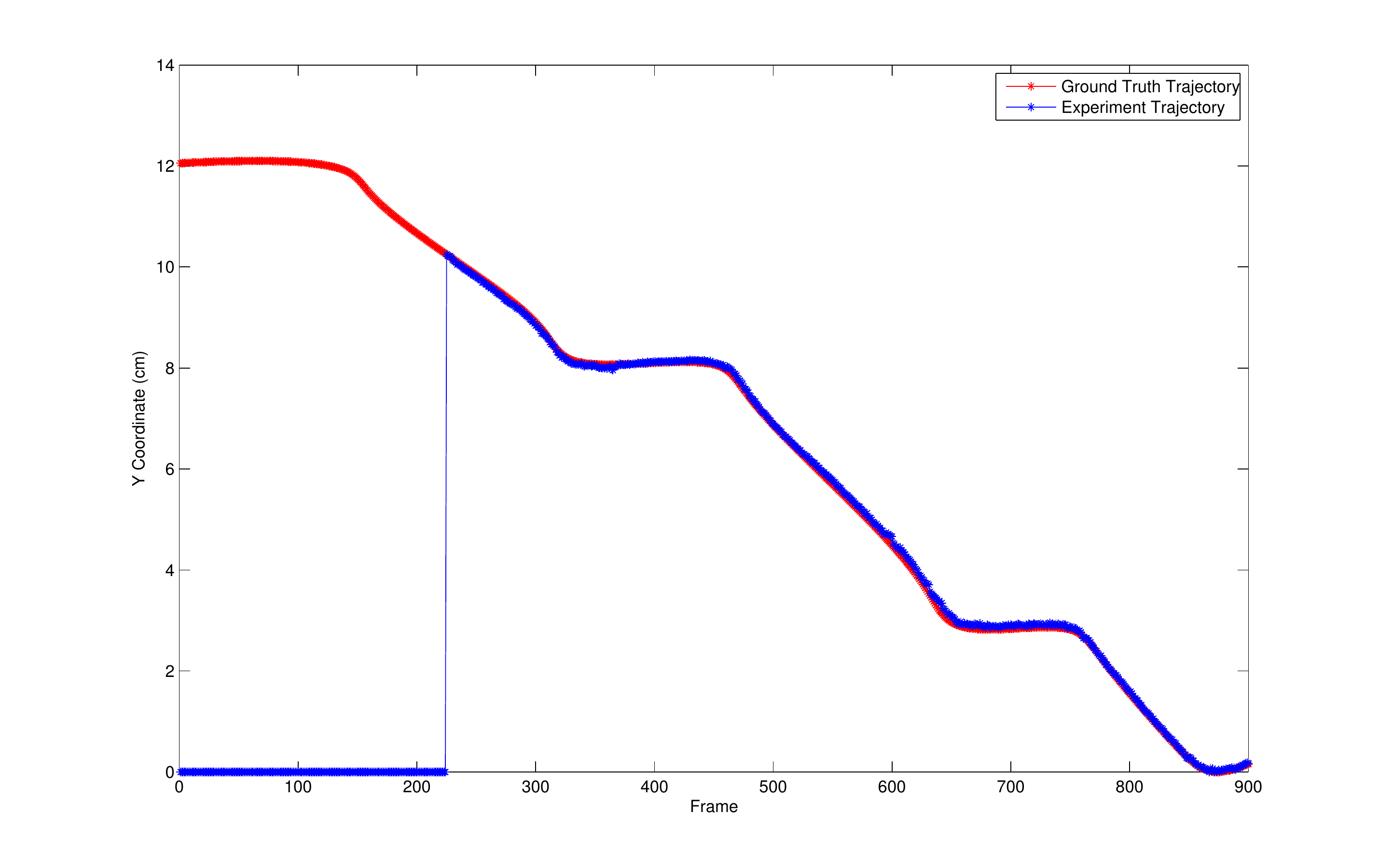}}\
\subfloat[]{\includegraphics[width=0.45\textwidth]{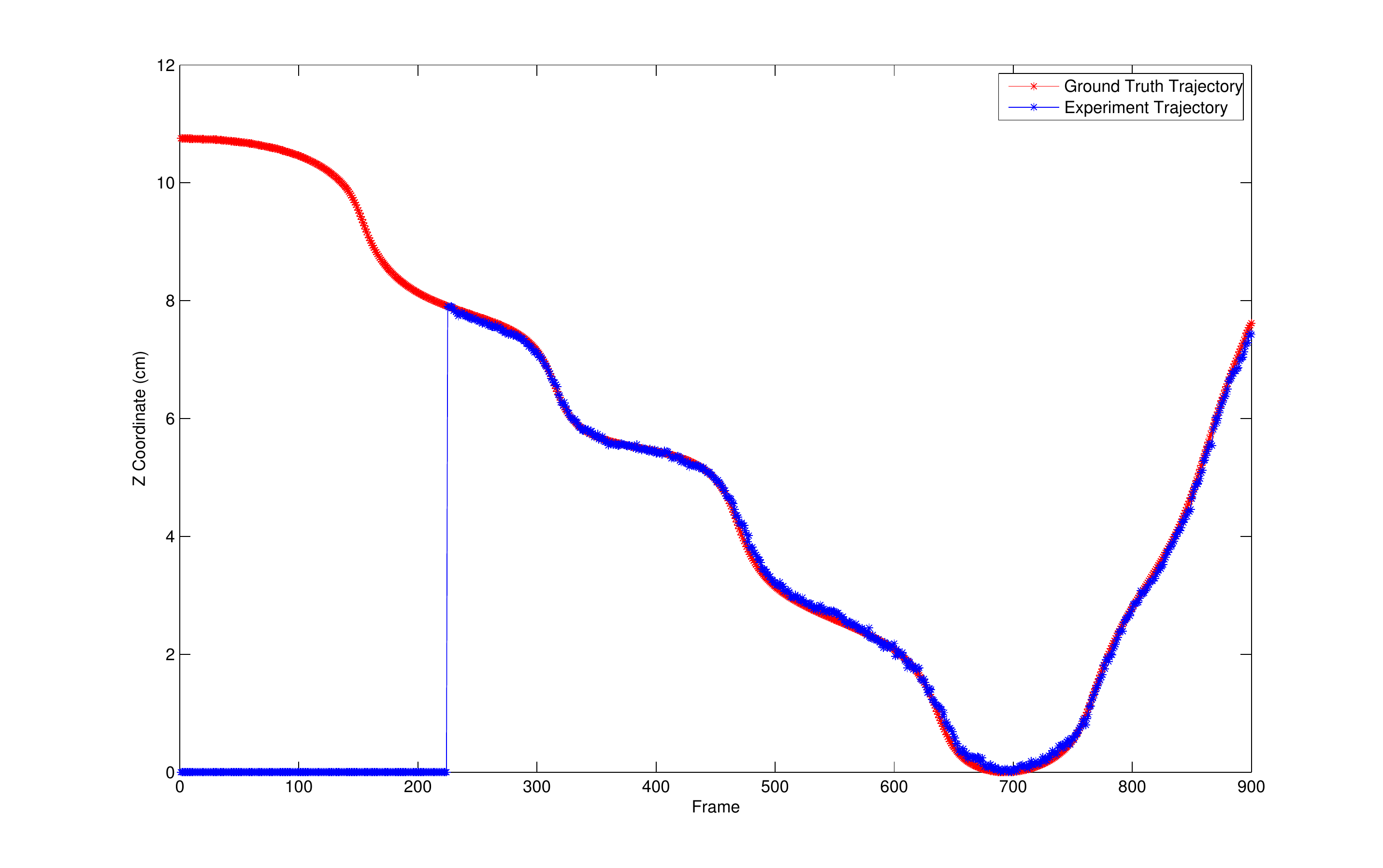}}\\
\caption{The camera trajectory comparison of the ground truth (red dots) with the estimated result (blue dots) in four different views, (a) 3D view, (b) view of X-axis, (c) view of Y-axis, (d) view of Z-axis}
\label{camera}       
\end{figure}

\subsubsection*{Camera trajectory evaluation}

Fig \ref{size}(c) shows one of the rendered images from the sequences used as the input to ORB--SLAM. The camera trajectory started with a close shot location to the liver surface. ORB--SLAM was successfully initialized at around frame 224 when the camera was in a place where many feature points were identified. After the initialization step, the SLAM system ran stably and estimated the camera trajectory with the origin of the coordinate system at the initialized position. The estimated camera trajectory was then extracted and normalized into the same coordinate system with that of the simulated ground truth model to assess the SLAM tracking performance.

Fig. \ref{camera} shows the performance evaluation results; Fig. \ref{camera}(a) displays both camera trajectories in 3D space, in which blue dots represent the camera trajectory estimated by ORB--SLAM, whereas red dots are the trajectory of the simulated ground truth. Figs. \ref{camera}(b), (c), and (d) are two camera trajectories in X-axis, Y-axis, and Z-axis views, respectively.
As can be seen, the SLAM camera trajectory starts at the frame of 224, as there is no estimated data before initialization. Once the camera tracking is initialised, the trajectory of the camera matches closely to the ground truth camera trajectory represented by red dots. RMSE between the two camera trajectory data sets was also calculated with a result of 1.24mm.

\subsubsection*{3D surface reconstruction evaluation}

\begin{figure}
\centering
\subfloat[]{\includegraphics[width=0.5\textwidth]{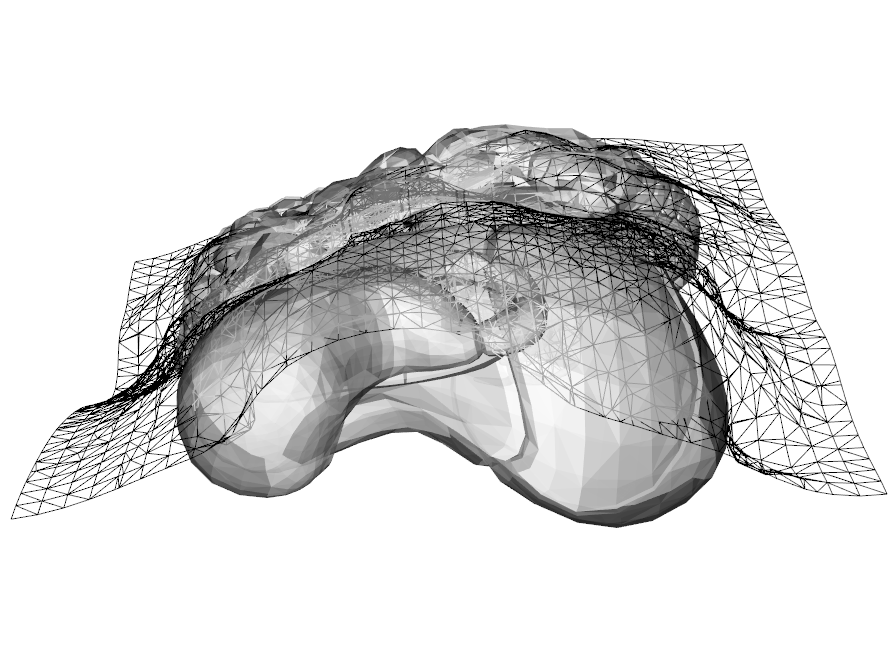}}\
\subfloat[]{\includegraphics[width=0.48\textwidth]{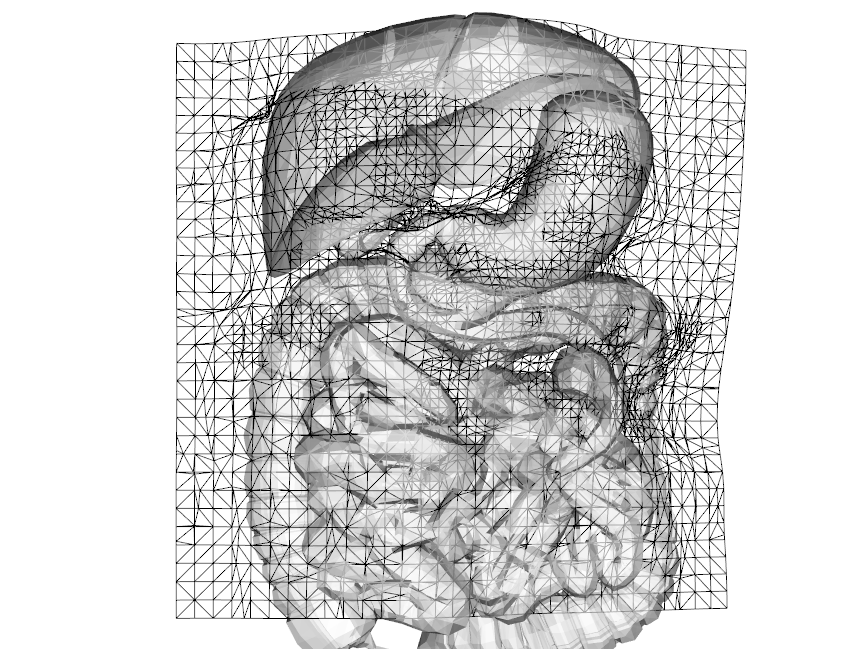}}\\
\subfloat[]{\includegraphics[width=0.45\textwidth]{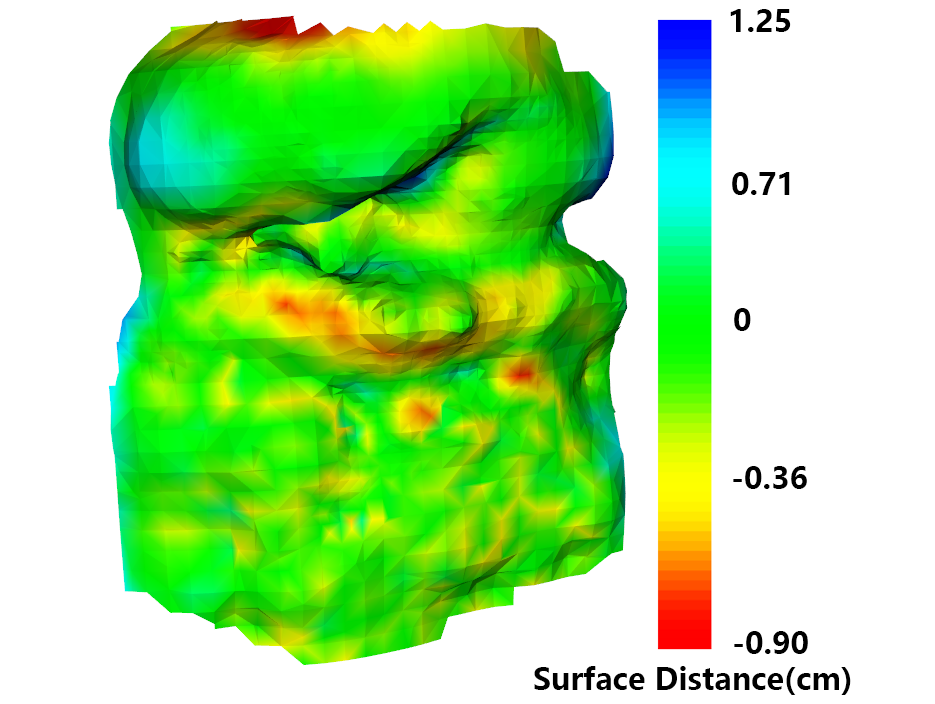}}\
\subfloat[]{\includegraphics[width=0.52\textwidth]{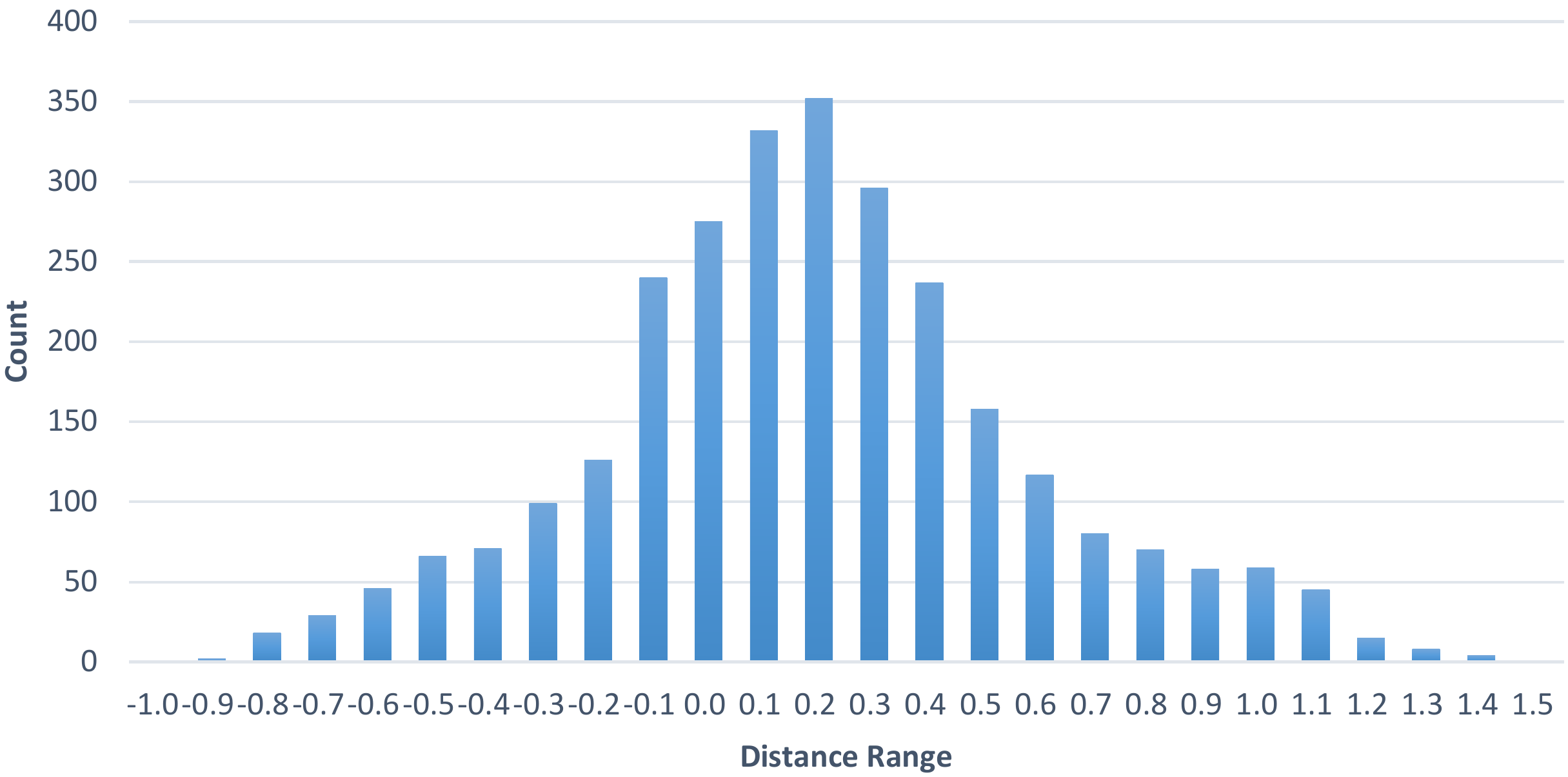}}\\
\caption{(a) and (b): the surface nicely represents the model surface. (c) Surface distance map between the reconstructed surface with the 3D model. (d)Distance distribution}
\label{surface}       
\end{figure}

When the ORB--SLAM system gained enough feature points, we build a 3D surface based on the sparse point cloud. The whole reconstruction pipeline takes only 600 ms to generate the surface, which was then exported into the 3D model space to be compared with the ground truth surface data set. A simple iterative closest point (ICP) algorithm was used to align the reconstructed surface with the 3D model that was used to render the video. 

Root Mean Squared Distance (RMSD) is used to evaluate the overall distance between the two surfaces. They are aligned in the real world coordinate system and we apply a grid sample to get a series of x,y coordinate points based on the surface area,  and then compare the distance of the z value of the two surfaces.
$$RMSD = \sqrt{\frac{1}{mn}\sum_{x=1}^{m}\sum_{y=1}^{n}\left ( Z_{x,y}-z_{x,y} \right )^{2}}$$
The RMSD to the ground truth surface is 4.32 mm for surface we reconstructed.

Fig. \ref{surface} (a) shows that the reconstructed 3D surface aligns with the 3D model closely; Fig. \ref{surface} (b) shows the top down view of the alignment. 
Fig. \ref{surface} (c) shows the distance map between the reconstructed surface with the 3D ground truth model, where warm colours show penetrations between the two surfaces, the green colour represents a perfect match between the two surfaces, and the blue colour shows the largest distance between the two surfaces.  Fig. \ref{surface} (d) illustrates the distance distribution, demonstrating a normal distribution with distances for most of the surface areas are between -1.0 mm to 4.0 mm. 

\subsection{Real endoscopic video evaluation}

\begin{figure}
\centering
\subfloat[]{\includegraphics[width=0.48\textwidth]{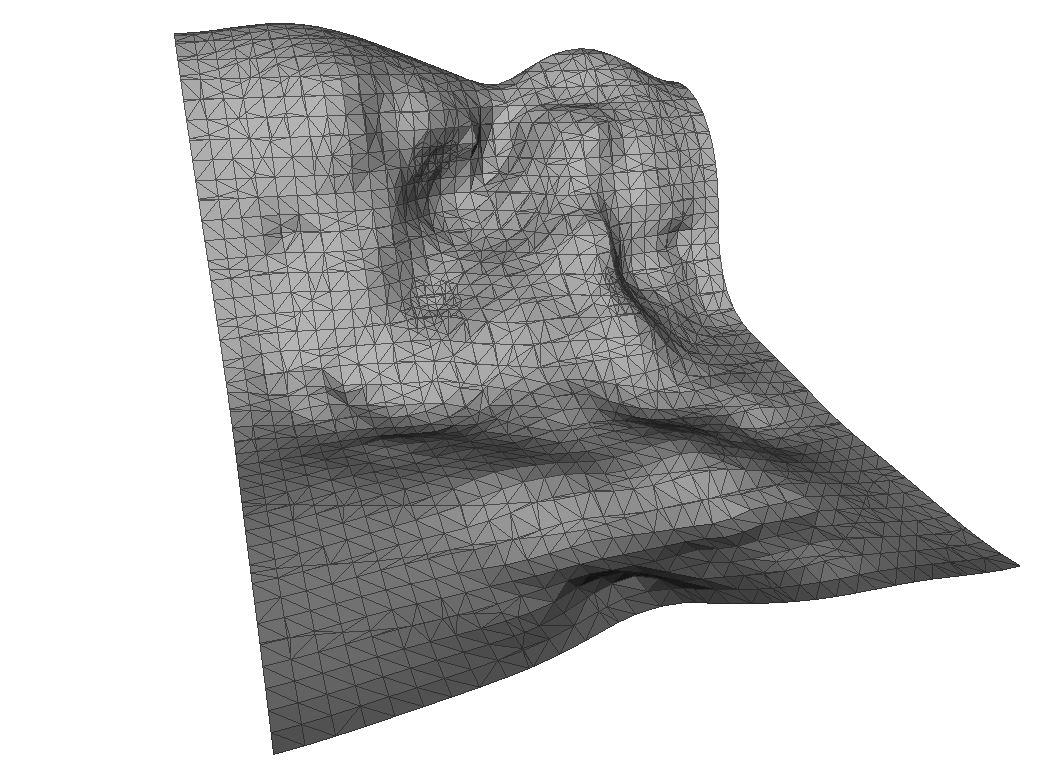}}\
\subfloat[]{\includegraphics[width=0.48\textwidth]{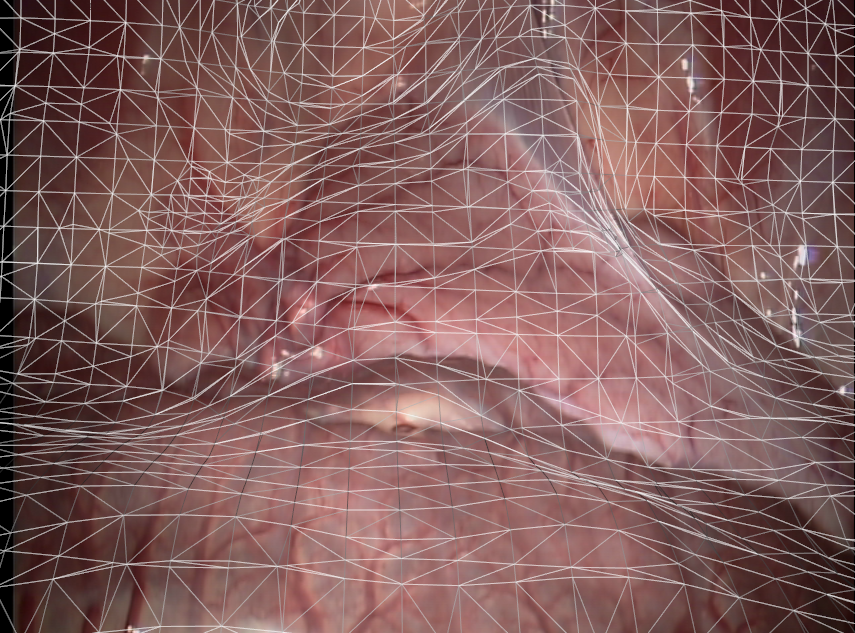}}\\
\subfloat[]{\includegraphics[width=0.48\textwidth]{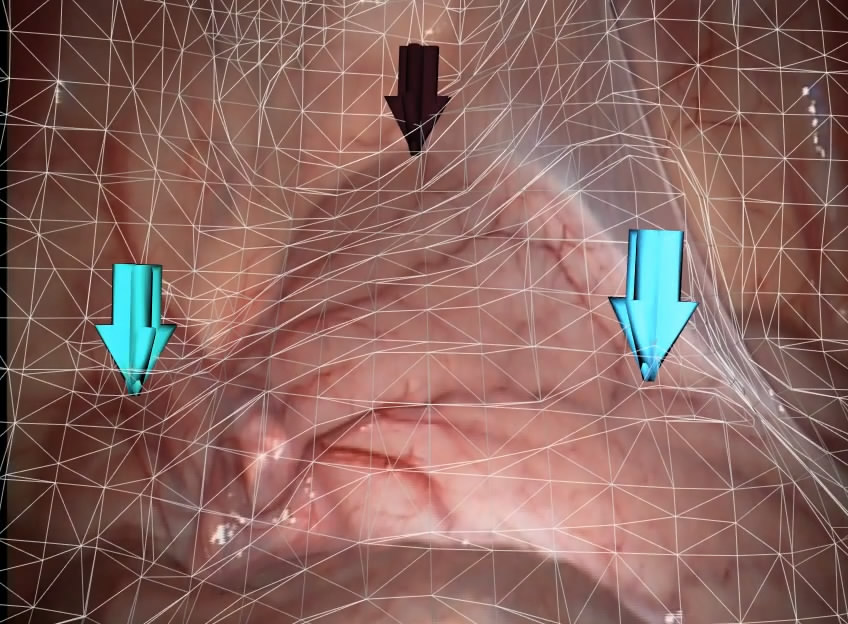}}\
\subfloat[]{\includegraphics[width=0.48\textwidth]{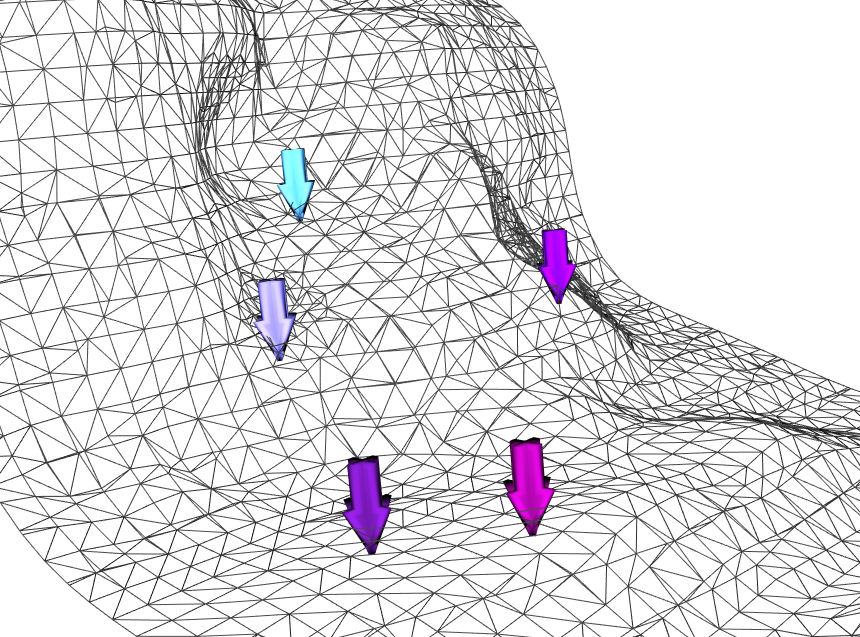}}\\
\caption{(a):The surface reconstruction results applied to an \textit{in vivo} video sequence. (b) The depth augmentation. (c) The AR element inserted with correct depth. (d) The mesh is manually rotated to show the depth.}
\label{real}       
\end{figure}

To qualitatively evaluate the performance of our proposed surface reconstruction framework, we applied the proposed approach on real \textit{in vivo} videos that we acquired from Hamlyn Centre Laparoscopic / Endoscopic Video Datasets \cite{London2016} \cite{Mountney2010}. Fig. \ref{real} (a) shows the reconstruction result from our 3D reconstruction framework. Fig. \ref{real} (b) shows the depth augmentation by fusing the camera pose from the SLAM system and the 3D surface reconstructed from our proposed framework. The real-time alignment of the 3D transparent mesh and the video augurs well for providing correct depth information intra-operatively and so help improve surgical performance by displaying 3D mesh structures when performing monocular endoscope procedures. 

Our new 3D surface reconstruction approach also allows us to develop a depth-correct AR framework for augmenting 3D models within the intra-operative endoscope scene in real-time. Depth-corrected AR is important when placing 3D models into the scene, since incorrect depth placement will cause virtual objects to appear to drift away when viewing perspective changes. Our depth-correct AR system ensures virtual objects can be superimposed at the correct positions. Fig.\ref{real} (c) shows some AR 3D-text notations placed into the video frames, and in Fig.\ref{real} (d), we manually rotated the mesh in order to inspect the depth of AR objects. More details can be appreciated in our video \cite{Youtube2016}.

\section{Conclusions}
\label{conclusions}
In this paper, we have proposed an efficient and effective 3D surface reconstruction framework for an intra-operative monocular laparosopic scene based on ORB--SLAM. This new approach has shown promising results when tested on both simulated laparosopic scene image sequences and clinical data. The proposed framework also augurs well for use with depth augmentation and augmented reality in MIS with correct depth. 

In future work, we will continue developing the dense SLAM system to be used in MIS reconstruction and extend the current reconstruction framework to get improved accuracy and speed. This will enable us to develop a prototype system that can be tested in the operating theatre with our clinical collaborators. This will further investigate the benefits and efficacy of this approach and provide evidence for our hypothesis that visual SLAM can enhance the tools available to the surgeon performing a monocular endoscopic procedure.

%


\bibliographystyle{spmpsci}      
\bibliography{bibtex}   

\begin{thebibliography}{10}
\providecommand{\url}[1]{{#1}}
\providecommand{\urlprefix}{URL }
\expandafter\ifx\csname urlstyle\endcsname\relax
  \providecommand{\doi}[1]{DOI~\discretionary{}{}{}#1}\else
  \providecommand{\doi}{DOI~\discretionary{}{}{}\begingroup
  \urlstyle{rm}\Url}\fi

\bibitem{Blender2016}
Blender: Blender - free and open 3d creation software (2016).
\newblock \urlprefix\url{https://www.blender.org/}.
\newblock [Accessed 6 Nov. 2016]

\bibitem{ChangStoyanovDavisonEtAl2013}
Chang, P.L., Stoyanov, D., Davison, A.J., Edwards, P.E.: Real-time dense stereo
  reconstruction using convex optimisation with a cost-volume for image-guided
  robotic surgery.
\newblock Med Image Comput Comput Assist Interv \textbf{16}(Pt 1), 42--49
  (2013)

\bibitem{Youtube2016}
Chen, L.: Youtube video (2016).
\newblock \urlprefix\url{https://youtu.be/m06dxtFeBOM}.
\newblock [Accessed 6 Nov. 2016]

\bibitem{GrasaBernalCasadoEtAl2014}
Grasa, O.G., Bernal, E., Casado, S., Gil, I., Montiel, J.: Visual slam for
  handheld monocular endoscope.
\newblock Medical Imaging, IEEE Transactions on \textbf{33}(1), 135--146 (2014)

\bibitem{HaouchineCotinPeterlikEtAl2015}
Haouchine, N., Cotin, S., Peterlik, I., Dequidt, J., Lopez, M.S., Kerrien, E.,
  Berger, M.O.: Impact of soft tissue heterogeneity on augmented reality for
  liver surgery.
\newblock Visualization and Computer Graphics, IEEE Transactions on
  \textbf{21}(5), 584--597 (2015)

\bibitem{HaouchineDequidtPeterlikEtAl2013}
Haouchine, N., Dequidt, J., Peterlik, I., Kerrien, E., Berger, M.O., Cotin, S.:
  Image-guided simulation of heterogeneous tissue deformation for augmented
  reality during hepatic surgery.
\newblock In: Mixed and Augmented Reality (ISMAR), 2013 IEEE International
  Symposium on, pp. 199--208. IEEE (2013)

\bibitem{Honeck2012}
Honeck, P., Wendt-Nordahl, G., Rassweiler, J., Knoll, T.: Three-dimensional
  laparoscopic imaging improves surgical performance on standardized ex-vivo
  laparoscopic tasks.
\newblock Journal of Endourology \textbf{26}(8), 1085--1088 (2012).
\newblock \doi{10.1089/end.2011.0670}.
\newblock \urlprefix\url{http://dx.doi.org/10.1089/end.2011.0670}

\bibitem{Kim2012}
Kim, J.H., Bartoli, A., Collins, T., Hartley, R.: Tracking by detection for
  interactive image augmentation in laparoscopy.
\newblock Lecture Notes in Computer Science pp. 246--255 (2012)

\bibitem{Kratzer2003}
Kratzer, W., Fritz, V., Mason, R.A., Haenle, M.M., Kaechele, V., ~, R.S.G.:
  Factors affecting liver size: a sonographic survey of 2080~{s}ubjects.
\newblock J Ultrasound Med \textbf{22}(11), 1155--1161 (2003)

\bibitem{Kumar2014}
Kumar, A., Wang, Y.Y., Wu, C.J., Liu, K.C., Wu, H.S.: Stereoscopic
  visualization of laparoscope image using depth information from {3D} model.
\newblock Computer Methods and Programs in Biomedicine \textbf{113}(3),
  862--868 (2014).
\newblock \doi{10.1016/j.cmpb.2013.12.013}.
\newblock \urlprefix\url{http://dx.doi.org/10.1016/j.cmpb.2013.12.013}

\bibitem{Levin2004}
Levin, D.: Mesh-independent surface interpolation.
\newblock Mathematics and Visualization pp. 37--49 (2004)

\bibitem{Lin2013}
Lin, B., Johnson, A., Qian, X., Sanchez, J., Sun, Y.: Simultaneous tracking,
  {3D} reconstruction and deforming point detection for stereoscope guided
  surgery.
\newblock Lecture Notes in Computer Science pp. 35--44 (2013)

\bibitem{London2016}
London, I.C.: Hamlyn centre laparoscopic / endoscopic video datasets (2016).
\newblock \urlprefix\url{http://hamlyn.doc.ic.ac.uk/vision/}.
\newblock [Accessed 6 Nov. 2016]

\bibitem{Michael2006}
Michael, K., Bolitho, M., Hoppe, H.: Poisson surface reconstruction.
\newblock In: Proceedings of the fourth Eurographics symposium on Geometry
  processing, vol.~7, p. 2006 (2006)

\bibitem{MountneyLoThiemjarusEtAl2007}
Mountney, P., Lo, B., Thiemjarus, S., Stoyanov, D., Zhong-Yang, G.: A
  probabilistic framework for tracking deformable soft tissue in minimally
  invasive surgery.
\newblock Med Image Comput Comput Assist Interv \textbf{10}(Pt 2), 34--41
  (2007)

\bibitem{Mountney2006}
Mountney, P., Stoyanov, D., Davison, A., Yang, G.Z.: Simultaneous stereoscope
  localization and soft-tissue mapping for minimal invasive surgery.
\newblock Med Image Comput Comput Assist Interv \textbf{9}(Pt 1), 347--354
  (2006)

\bibitem{Mountney2010}
Mountney, P., Stoyanov, D., Yang, G.Z.: Three-dimensional tissue deformation
  recovery and tracking.
\newblock IEEE Signal Processing Magazine \textbf{27}(4), 14--24 (2010).
\newblock \doi{10.1109/msp.2010.936728}.
\newblock \urlprefix\url{http://dx.doi.org/10.1109/MSP.2010.936728}

\bibitem{MountneyYang2008}
Mountney, P., Yang, G.Z.: Soft tissue tracking for minimally invasive surgery:
  learning local deformation online.
\newblock Med Image Comput Comput Assist Interv \textbf{11}(Pt 2), 364--372
  (2008)

\bibitem{Mountney2009}
Mountney, P., Yang, G.Z.: Dynamic view expansion for minimally invasive surgery
  using simultaneous localization and mapping.
\newblock 2009 Annual International Conference of the IEEE Engineering in
  Medicine and Biology Society  (2009).
\newblock \doi{10.1109/iembs.2009.5333939}.
\newblock \urlprefix\url{http://dx.doi.org/10.1109/IEMBS.2009.5333939}

\bibitem{MountneyYang2010}
Mountney, P., Yang, G.Z.: Motion compensated slam for image guided surgery.
\newblock Medical Image Computing and Computer-Assisted Intervention MICCAI
  2010  (2010)

\bibitem{Mur-ArtalMontielTardos2015}
Mur-Artal, R., Montiel, J.M.M., Tard{\'{o}}s, J.D.: Orb-{SLAM}: A versatile and
  accurate monocular {SLAM} system.
\newblock IEEE Transactions on Robotics \textbf{31}(5), 114--1163 (2015).
\newblock \doi{10.1109/TRO.2015.2463671}

\bibitem{NaderMahmoud2016}
Nader~Mahmoud Inigo~Cirauqui, A.H.C.D.L.S.J.M.J.M.: Orbslam-based endoscope
  tracking and 3d reconstruction.
\newblock In: MICCAI 2016 CARE (2016)

\bibitem{Plantefeve2016}
Plantef{\'{e}}ve, R., Peterlik, I., Haouchine, N., Cotin, S.: Patient-specific
  biomechanical modeling for guidance during minimally-invasive hepatic
  surgery.
\newblock Ann Biomed Eng \textbf{44}(1), 139--153 (2016).
\newblock \doi{10.1007/s10439-015-1419-z}.
\newblock \urlprefix\url{http://dx.doi.org/10.1007/s10439-015-1419-z}

\bibitem{StoyanovDarziYang2004}
Stoyanov, D., Darzi, A., Yang, G.Z.: Dense 3d depth recovery for soft tissue
  deformation during robotically assisted laparoscopic surgery.
\newblock Medical Image Computing and Computer-Assisted Intervention MICCAI
  2004  (2004)

\bibitem{Stoyanov2005}
Stoyanov, D., Darzi, A., Yang, G.Z.: A practical approach towards accurate
  dense 3d depth recovery for robotic laparoscopic surgery.
\newblock Comput Aided Surg \textbf{10}(4), 199--208 (2005).
\newblock \doi{10.3109/10929080500230379}.
\newblock \urlprefix\url{http://dx.doi.org/10.3109/10929080500230379}

\bibitem{StoyanovScarzanellaPrattEtAl2010}
Stoyanov, D., Scarzanella, M.V., Pratt, P., Yang, G.Z.: Real-time stereo
  reconstruction in robotically assisted minimally invasive surgery.
\newblock Medical Image Computing and Computer-Assisted Intervention MICCAI
  2010  (2010)

\bibitem{Totz2011}
Totz, J., Mountney, P., Stoyanov, D., Yang, G.Z.: Dense surface reconstruction
  for enhanced navigation in mis.
\newblock Med Image Comput Comput Assist Interv \textbf{14}(Pt 1), 89--96
  (2011)

\bibitem{Wagner2012}
Wagner, O.J., Hagen, M., Kurmann, A., Horgan, S., Candinas, D., Vorburger,
  S.A.: Three-dimensional vision enhances task performance independently of the
  surgical method.
\newblock Surg Endosc \textbf{26}(10), 2961--2968 (2012).
\newblock \doi{10.1007/s00464-012-2295-3}.
\newblock \urlprefix\url{http://dx.doi.org/10.1007/s00464-012-2295-3}

\bibitem{Ye2016}
Ye, M., Giannarou, S., Meining, A., Yang, G.Z.: Online tracking and retargeting
  with applications to optical biopsy in gastrointestinal endoscopic
  examinations.
\newblock Medical Image Analysis \textbf{30}, 14--157 (2016).
\newblock \doi{10.1016/j.media.2015.10.003}.
\newblock \urlprefix\url{http://dx.doi.org/10.1016/j.media.2015.10.003}

\end{thebibliography}

\end{document}